\documentclass[10pt,a4paper]{article}

\usepackage[utf8]{inputenc}
\usepackage[T1]{fontenc}
\usepackage{amsmath,amssymb,amsfonts}
\usepackage{graphicx}
\usepackage{hyperref}
\usepackage{algorithm}
\usepackage{algpseudocode}
\usepackage{multicol}
\usepackage{geometry}
\usepackage{cite}
\usepackage{authblk}
\usepackage{amsmath, mathtools, empheq}

\title{\Large\textbf{TPTT: Transforming Pretrained Transformers into Titans}}
\author{\large Fabien Furfaro\thanks{\texttt{fabien.furfaro@gmail.com}}}
\date{\large 2025}

\begin{document}

\maketitle
    
\begin{abstract}
    \textbf{Background:} Transformer-based large language models (LLMs) have achieved strong performance across many natural language processing tasks. Nonetheless, their quadratic computational and memory requirements, particularly in self-attention layers, pose challenges for efficient inference on long contexts and for deployment in resource-limited environments.

    \textbf{Methods:} We present TPTT (Transforming Pretrained Transformers into Titans), a framework designed to augment pretrained Transformers with linearized attention and internal memory gating via Memory as Gate (MaG), applied without full retraining. TPTT supports parameter-efficient fine-tuning (LoRA) and integrates with standard toolkits such as Hugging Face Transformers. We evaluated TPTT on several pretrained models, including Llama-1B, OlMoE-1B-7B, Qwen2.5-1.5B, Gemma3-270m, OpenELM-1.3B, and Mistral-7B, in order to assess applicability across architectures of different scales.

    \textbf{Results:} Experiments on models with approximately 1 billion parameters, evaluated primarily on the MMLU benchmark, suggest potential improvements in both efficiency and accuracy compared to baseline models. For example, Titans-Llama-1B exhibited up to a 20\% relative increase in Exact Match scores in one-shot evaluation. An additional finding is that it is possible to convert a quadratic-attention model into a purely linear-attention model using the DeltaProduct mechanism. All training runs were carried out with modest computational resources.

    \textbf{Conclusions:} These preliminary findings indicate that TPTT may help adapt pretrained LLMs for long-context tasks with limited overhead. Further studies on larger models and a broader set of benchmarks will be necessary to evaluate the generality and robustness of the framework. Source code and package are openly available at \url{https://github.com/fabienfrfr/tptt} and \url{https://pypi.org/project/tptt/}.
\end{abstract}

\begin{figure}[ht]
    \centering
    \includegraphics[width=0.8\linewidth]{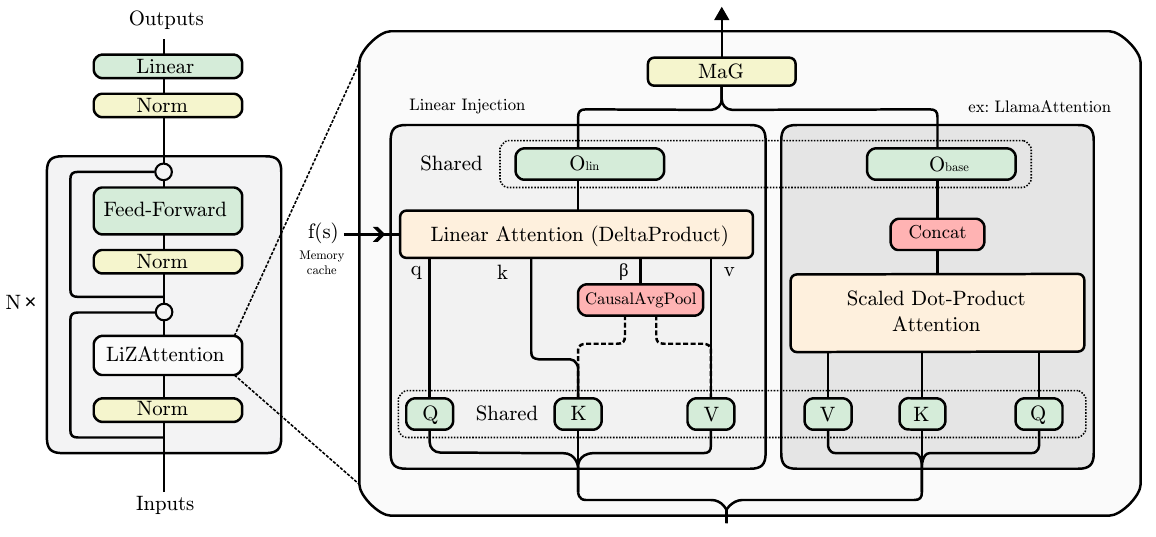}
    \caption{Overview of the TPTT architecture. On the left, the diagram illustrates a decoder-only architecture where linear attention is injected in parallel of vanilla attention (LiZAttention). On the right, the detailed architecture of the linearized attention mechanism is depicted, highlighting the shared weights for query (Q), key (K), value (V), and output (O) projections. It also shows the management of the state memory (S) and the combination of outputs through the Memory as Gate (MaG) weighting mechanism. The state can be unlinear between chunk. The diagram emphasizes the integration of linearized attention mechanisms and advanced memory management techniques, such as DeltaProduct and CausalAvgPool1D, contributing to processing and output generation.}
    \label{fig:approach_overview}
\end{figure}

\section{Introduction}

Transformer-based large language models (LLMs) have revolutionized natural language processing (NLP), achieving state-of-the-art results across a wide range of tasks, from text generation to code synthesis~\cite{vaswani2017attention, mann2020language}. However, the quadratic complexity of the self-attention mechanism in both computation and memory remains a major limitation, particularly for applications requiring long-context inference, such as document understanding, multi-turn dialogue, and scientific reasoning.

Recent research has sought to address these challenges through a variety of approaches. Efficient attention mechanisms~\cite{katharopoulos2020transformers, yang2024parallelizing} aim to reduce the computational burden by approximating or restructuring the attention operation, while recurrent and memory-augmented architectures~\cite{mercat2024linearizing, behrouz2024titans} introduce internal memory to better capture long-range dependencies. At the same time, parameter-efficient fine-tuning techniques such as LoRA~\cite{hu2022lora} have made it feasible to adapt large pretrained models to new domains or tasks without the need for full retraining.

Despite these advances, most existing solutions require substantial architectural changes or retraining from scratch, limiting their applicability to already deployed or proprietary models. There remains a need for methods that can efficiently endow existing LLMs with long-context and memory-augmented capabilities, while preserving their pretrained knowledge and minimizing adaptation cost.

\textbf{In this paper}, we introduce TPTT (Transforming Pretrained Transformers into Titans), a practical and scalable framework for upgrading pretrained Transformers with efficient linearized attention and advanced memory mechanisms. Our approach integrates techniques such as Memory as Gate (MaG) and mixed linearized attention (LiZA), inspired by recent advances in memory-augmented neural architectures~\cite{behrouz2024titans, behrouz2025s}. TPTT is fully compatible with the Hugging Face Transformers library and leverages parameter-efficient fine-tuning (LoRA), enabling seamless adaptation of any causal LLM to long-context tasks (see Figure~\ref{fig:approach_overview}).

\textbf{Our main contributions are:}

\begin{itemize}
    \item We propose TPTT, a framework for augmenting pretrained Transformers with linearized attention and memory gating, requiring minimal changes to the original architecture.
    \item We introduce the MaG mechanism, which adaptively combines linear and softmax attention outputs, and demonstrate its effectiveness for long-context modeling.
    \item We provide an open-source implementation compatible with popular frameworks, and show that TPTT enables substantial improvements in both efficiency and accuracy on challenging benchmarks.
\end{itemize}

\section{Related Work}

The quadratic cost of self-attention in Transformers has motivated extensive research on efficiency and scalability. Prior work broadly falls into four main directions.

\paragraph{Attention Mechanisms}  
Implementation-level optimizations such as FlashAttention~\cite{dao2023flashattention} improve runtime efficiency and reduce memory footprint, often by fusing kernels, introducing tiling strategies, or developing custom CUDA implementations. These techniques yield practical speedups for training and inference but do not alter the quadratic scaling of the standard attention mechanism with respect to sequence length.

\paragraph{Linearized and Approximated Attention}  
Several methods reformulate attention into linear or low-rank approximations. Linformer~\cite{wang2020linformer} compresses sequence representations via low-rank projections, while Performer~\cite{choromanski2020rethinking} employs random feature maps for kernel-based approximations.  Mamba replaces standard attention altogether with a selective state space model (SSM) ~\cite{dao2024transformers}. More recent work on chunkwise-parallel formulations, such as Parallelizing Linear Attention~\cite{yang2024parallelizing}, further improves the practicality of linear attention at scale. DeltaProduct~\cite{siems2025deltaproduct} belongs to this family, using products of generalized Householder transformations to enhance expressivity while preserving stability. These approaches reduce the computational bottleneck but typically require models to be trained from scratch or with significant architectural modifications.

\paragraph{Recurrent and Memory-Augmented Architectures}  
Alternative designs replace or complement self-attention with recurrence or internal memory. RWKV~\cite{peng2023rwkv} adopts a recurrent formulation to capture dependencies in a sequence-efficient way, while Titans~\cite{behrouz2024titans} introduce long-range memory-augmented mechanisms. Notably, DeltaProduct achieves a level of expressivity comparable to Titans~\cite{behrouz2025s}, but both methods generally require new models to be trained rather than adaptation of existing ones.

\paragraph{Parameter-Efficient Fine-Tuning}  
Methods such as LoRA~\cite{hu2022lora} and adapters reduce the computational cost of fine-tuning by introducing small trainable components while freezing the majority of pretrained parameters. Recent work has extended this idea, such as LoLCat~\cite{zhang2024lolcats} and Liger~\cite{lan2025liger}, which integrate low-rank or structured adaptations into fine-tuning pipelines to further improve efficiency. However, the integration of efficient attention or memory mechanisms via parameter-efficient tuning remains relatively underexplored. For example, Liger combines gated linear attention with lightweight adaptation, offering efficiency but potentially reduced expressivity compared to mechanisms such as DeltaProduct~\cite{siems2025deltaproduct}.

\paragraph{Our Positioning}  
TPTT addresses a complementary gap. Rather than designing entirely new architectures or relying solely on low-level optimizations, it provides a framework to retrofit pretrained Transformers with linearized attention and memory gating, requiring minimal architectural changes. Combined with parameter-efficient tuning (e.g. LoRA), this approach may offer a practical path to extend the capabilities of existing large language models to longer contexts without full retraining.

\section{Methodology}

\subsection{Linearized Attention Mechanisms}

Standard self-attention in transformers computes pairwise interactions between all tokens, resulting in quadratic complexity with respect to sequence length (i.e., $O(T^2 D)$ where $T$ is the sequence length and $D$ the feature dimension)~\cite{vaswani2017attention}. To address this computational bottleneck, linearized attention mechanisms approximate the softmax attention by projecting queries and keys through a carefully chosen feature map $\phi$~\cite{katharopoulos2020transformers, wang2020linformer, mercat2024linearizing, yang2024parallelizing}. Typical choices for $\phi$ include kernel feature maps or nonlinearities such as ELU, which enable rewriting the attention as a series of linear operations. This approximation reduces both computational and memory costs, allowing efficient processing of long sequences while maintaining substantial modeling power~\cite{gu2023mamba, dao2023flashattention, zhang2024lolcats, lan2025liger}.

However, linearized attention approaches can suffer from reduced expressivity compared to full softmax attention, especially in capturing complex, non-linear dependencies across tokens. This motivates the development of mechanisms that balance efficiency and expressivity.

Formally, given an input sequence $\mathbf{X} = \{\mathbf{x}_1, \dots, \mathbf{x}_T\} \in \mathbb{R}^{T \times D}$, the standard self-attention computes:
\begin{equation}
\mathbf{Q}, \mathbf{K}, \mathbf{V} = \mathbf{X}\mathbf{W}_Q, \mathbf{X}\mathbf{W}_K, \mathbf{X}\mathbf{W}_V
\end{equation}
\begin{equation}
\mathbf{O}_{\text{base}} = \text{Softmax}\left(\frac{\mathbf{Q}\mathbf{K}^\top}{\sqrt{D}}\right)\mathbf{V}
\end{equation}

In linear attention, the softmax kernel is approximated by a feature mapping $\phi$, yielding the output at position $t$:
\begin{equation}
\mathbf{O}_{\text{lin}, (t)} = 
\frac{
\phi(\mathbf{q}_t)^\top \left( \sum_{i=1}^t \phi(\mathbf{k}_i) \beta_i\, \mathbf{v}_i^\top \right)}{
\phi(\mathbf{q}_t)^\top \left( \sum_{i=1}^t \phi(\mathbf{k}_i) \beta_i \right)} 
= \frac{\phi(\mathbf{q}_t)^\top S_t}{\phi(\mathbf{q}_t)^\top z_t}
\end{equation}
where $\mathbf{q}_t$, $\mathbf{k}_i$, and $\mathbf{v}_i$ are the query, key, and value vectors at positions $t$ and $i$, respectively, and $\beta_i$ is a gating scalar or vector modulating the keys and values. Here, $\beta_i$ is applied multiplicatively after the feature mapping $\phi$, which itself can be an ELU-based or kernel approximation function~\cite{wang2020linformer}. In this work, the projected queries, keys, and values are normalized with 
the SiLU activation, the gating coefficients $\beta_i$ are obtained via causal average pooling with a sigmoid, and the attention output is normalized with RMSNorm for stability:

\begin{align}
    \{q_i,k_i,v_i\} &\;\leftarrow \frac{\mathrm{SiLU}(\{q_i,k_i,v_i\})}{\|\mathrm{SiLU}(\{q_i,k_i,v_i\})\|_2 + \epsilon}, \\
    \{\beta_k, \beta_v\} &\;=\; \sigma\!\big(\mathrm{Pool}_{1:i}(\{k, v\})\big), \\
    \mathbf{O}_{\text{lin}, (t)}  &\;=\; \mathrm{RMSNorm}(\mathbf{y}) .
\end{align}

Here $\{q_i,k_i,v_i\}$ denotes the normalized queries, keys, and values; $\{\beta_k, \beta_v\}$ the gating vector obtained from a \emph{causal} average pooling over past keys and values (i.e., $1{:}i$); and $\mathbf{y}$ the delta product attention output.


\subsection{Memory as Gate (MaG)}
To further improve modeling of long-range dependencies, we introduce \textit{Memory as Gate} (MaG), an internal memory augmentation mechanism inspired by Titans~\cite{behrouz2024titans}. MaG enables the model to dynamically combine the strengths of linearized and softmax attention, adapting to the context and sequence characteristics. Given the outputs of linearized attention $\mathbf{O}_{\text{lin}}$ and softmax attention $\mathbf{O}_{\text{base}}$, MaG computes the final attention output through two possible strategies:
\begin{enumerate}
    \item \textbf{Gated Mixing:} A learnable parameter $\alpha$ is broadcast to match the attention tensor shapes. The outputs are combined as:
    \begin{equation}
        \mathbf{O}_{\text{MaG}} = (1-\alpha) \cdot \mathbf{O}_{\text{base}} + \alpha \cdot \mathbf{O}_{\text{lin}} .
    \end{equation}
    \item \textbf{Cross-Gate Mixing:} Optionally, MaG can introduce a nonlinear interaction term between the two attention outputs:
    \begin{equation}
        \mathbf{O}_{\text{MaG}} = (1-\alpha) \cdot \mathbf{O}_{\text{base}} + \alpha \cdot \mathbf{O}_{\text{lin}} + \left[(1-\alpha) \cdot \mathbf{O}_{\text{base}}\right] \odot \left[\alpha \cdot \mathbf{O}_{\text{lin}}\right] .
    \end{equation}
\end{enumerate}

Here, $\alpha \in [0,1]$ controls the mixing ratio between softmax and linearized attention, and $\odot$ denotes elementwise multiplication. While $\alpha$ could in principle be defined as a learnable parameter (e.g., per-layer or per-head), in our experiments it was not optimized by backpropagation. Instead, $\alpha$ was scheduled externally during training via the LiZA callback (see Sec.~\ref{sec:training}). This design avoids introducing new trainable parameters while still allowing the model to progressively adjust the balance between efficiency and expressivity. The gating mechanism can thus be interpreted as a form of memory-augmented control, where the reliance on linearized versus softmax attention evolves according to the training schedule and input context.

\subsection{Parallel Delta Product Modeling}

We leverage the \textbf{DeltaProduct} operator as the feature mapping $\phi$ within the linear attention framework, generalizing the classical Delta rule to increase expressivity while preserving efficient parallel computation. DeltaProduct addresses the expressivity bottleneck found in linear recurrent architectures that rely on diagonal or diagonal-plus-rank-1 state transitions by employing a product of multiple generalized Householder transformations per token~\cite{siems2025deltaproduct}. This structure also ensures numerical stability due to the orthogonality properties of Householder matrices.

\subsubsection{Delta Rule Formulation}

\paragraph{Sequential Delta Rule.}

The classical Delta rule for associative memory updates the state matrix $\mathbf{S}_t \in \mathbb{R}^{d \times d}$ at time $t$ as:
\begin{align}
    \mathbf{S}_t &= \mathbf{S}_{t-1} \left(\mathbf{I}  - \beta_t \mathbf{k}_t  \mathbf{k}_t^\top \right) + \beta_t \mathbf{v}_t \mathbf{k}_t^\top \\
    &= \mathbf{S}_{t-1} + \beta_t \left( \mathbf{v}_t - \mathbf{S}_{t-1} \mathbf{k}_t \right) \mathbf{k}_t^\top
    \label{eq:deltarule}
\end{align}
where $\mathbf{k}_t, \mathbf{v}_t \in \mathbb{R}^d$ are the key and value vectors, and $\beta_t$ is a gating scalar. This update is inherently sequential and thus not well-suited for parallel hardware acceleration.

\paragraph{Expanded recursive form.}

Expanding the recursion, the state $\mathbf{S}_t$ can be expressed as the cumulative effect of all past updates modulated by subsequent corrections, i.e.,
\begin{equation}
    \mathbf{S}_t = \sum_{i=1}^t \beta_i \mathbf{v}_i \mathbf{k}_i^\top \prod_{j=i+1}^t \left( \mathbf{I} - \beta_j \mathbf{k}_j \mathbf{k}_j^\top \right),
\end{equation}
which highlights how each rank-1 update is successively transformed by the multiplicative decay terms from later steps.

\paragraph{Chunkwise Parallelization.}

To enable parallel computation, the input sequence is partitioned into chunks of size $C$. The update within each chunk is computed in parallel as~\cite{yang2024parallelizing}:
\begin{align}
    \mathbf{S}_{[t,\,t+C-1]} &= \mathbf{S}_{t-1} + \mathbf{K}^\top \mathbf{Y} \\
    \mathbf{Y} &= \mathbf{T}^{-1} \mathbf{R} \\
    \mathbf{R} &= \mathbf{V} \odot \boldsymbol{\beta} - (\mathbf{K} \odot \boldsymbol{\beta}) \mathbf{S}_{t-1} \\ 
    \mathbf{T} &= \mathbf{I} - \mathrm{tril}\left((\mathbf{K} \odot \boldsymbol{\beta}) \mathbf{K}^\top, -1\right)
\end{align}
where $\mathbf{K}, \mathbf{V} \in \mathbb{R}^{C \times d}$ are stacked keys and values for the chunk, $\boldsymbol{\beta} \in \mathbb{R}^{C \times 1}$ is the gating vector, $\odot$ denotes elementwise multiplication, and with $\mathrm{tril}(\cdot, -1)$ extracting the strictly lower triangular part. This formulation enables efficient batched computation by restricting dependencies within chunks.

\subsubsection{Delta Product}

\paragraph{DeltaProduct: Higher-Order Parallel State Updates.}

DeltaProduct generalizes the above by applying $n_h$ rank-1 Householder-like updates per token. The state update at position $i$ is recursively defined as:
\begin{equation}
    \mathbf{H}_{i, j} = \mathbf{H}_{i, j-1} + \beta_{i, j} \left( \mathbf{v}_{i, j} - \mathbf{H}_{i, j-1} \mathbf{k}_{i, j} \right) \mathbf{k}_{i, j}^\top, \quad j=1,\ldots,n_h
    \label{eq:deltaproduct-recurrence}
\end{equation}
with $\mathbf{H}_{i, 0} = \mathbf{H}_{i-1}$ and $\mathbf{H}_{i, n_h} = \mathbf{H}_i$. Each $\mathbf{k}_{i, j}, \mathbf{v}_{i, j}, \beta_{i, j}$ is generated from the input at position $i$ for the $j$-th Householder step, typically via independent learned linear projections.

\paragraph{Expanded recursive form.}

By unrolling the recursive updates over the internal steps \(j = 1, \ldots, n_h\) for token \(i\), the state \(\mathbf{H}_i\) can be equivalently expressed as:

\begin{equation}
\mathbf{H}_i = \prod_{j=1}^{n_h} \left(\mathbf{I} - \beta_{i,j} \mathbf{k}_{i,j} \mathbf{k}_{i,j}^\top \right) \mathbf{H}_{i-1} 
+ \sum_{j=1}^{n_h} \left( \beta_{i,j} \mathbf{k}_{i,j} \mathbf{v}_{i,j}^\top \prod_{\ell = j+1}^{n_h} \left(\mathbf{I} - \beta_{i,\ell} \mathbf{k}_{i,\ell} \mathbf{k}_{i,\ell}^\top \right) \right)
\end{equation}

where:
- The product \(\prod_{j=1}^{n_h} (\cdot)\) denotes the ordered matrix product (applied from right to left).
- The first term corresponds to the state transition matrix obtained by chaining \(n_h\) generalized Householder reflections applied to the previous token’s state \(\mathbf{H}_{i-1}\).
- The second term sums the rank-1 corrections given by the values \(\mathbf{v}_{i,j}\), each transformed by subsequent Householder matrices.
  
This expansion reveals that DeltaProduct’s transition matrix is a product of $n_h$ such transformations, interpolating between simple DeltaNet (where $n_h=1$) and richer, more expressive state transitions. It explicitly encodes how multiple gradient steps per token compose to update the memory state, with each rank-1 update being successively modulated by multiplicative decay terms from later steps. \footnote{\textit{Intuitively, this means that the memory at time $t$ is not a raw sum of past contributions, but rather a history filtered through successive decays, giving more weight to recent information while progressively fading distant dependencies.}}

\paragraph{Chunkwise DeltaProduct Update.}

For a chunk of $C$ real tokens (expanded to $N = n_h C$ virtual tokens), the parallel state update is:
\begin{align}
    \mathbf{T}_n &= \mathbf{I}_N - \mathrm{tril}\left((\mathbf{K}_n \odot \boldsymbol{\beta}_n) \mathbf{K}_n^\top, -1\right) \label{eq:triangular-nh} \\
    \mathbf{U}_n &= \mathbf{V}_n \odot \boldsymbol{\beta}_n - (\mathbf{K}_n \odot \boldsymbol{\beta}_n) \mathbf{S}_{\mathrm{in}} \\
    \mathbf{y}_n &= \mathbf{T}_n^{-1} \mathbf{U}_n \\
    \mathbf{S}_{[t,\,t+C-1]} &= \mathbf{S}_{\mathrm{in}} + \mathbf{K}_n^\top \mathbf{y}_n
    \label{eq:chunk-deltaproduct}
\end{align}
where $\mathbf{K}_n, \mathbf{V}_n \in \mathbb{R}^{N \times d}$ and $\boldsymbol{\beta}_n \in \mathbb{R}^{N \times 1}$ are the keys, values, and gates for all virtual tokens in the chunk, and $\mathbf{S}_{\mathrm{in}}$ is the initial state for the chunk (typically the last state of the previous chunk).

\paragraph{Virtual Token Expansion: Motivation and Formulation.}

To enable multiple Householder updates per token in parallel, each real token $\mathbf{x}_t \in \mathbb{R}^D$ is expanded into $n_h$ virtual tokens using strategies designed to enrich expressivity while maintaining numerical stability. We have two types of transformations:

\begin{equation}
    \text{Derivative trick:} \quad \mathbf{x}_{t,\mathrm{deriv}}^{(m)} = \frac{1}{Z_m} \sum_{k=0}^m (-1)^k \binom{m}{k} \mathbf{x}_{t-k}, \quad m=0,\ldots,n_h - 1,
    \label{eq:derivative-trick}
\end{equation}

\begin{equation}
    \text{Rotary trick:} \quad \mathbf{x}_{t,\mathrm{rot}}^{(m)} = \mathbf{R}(\theta_m) \mathbf{x}_t, \quad \text{with} \quad \theta_m = \frac{2\pi m}{n_h},
    \label{eq:rotative-trick}
\end{equation}

where $\mathbf{R}(\theta_m)$ applies a phase rotation of angle $\theta_m$ to consecutive pairs of features of $\mathbf{x}_t$, and $Z_m$ normalizes the scale. This expansion is applied independently to each input sequence $\mathbf{Q}$, $\mathbf{K}$, $\mathbf{V}$, and $\boldsymbol{\beta}$, resulting in $N = n_h C$ virtual tokens per chunk.

\paragraph{Nonlinearity.}

Optionally, a nonlinearity $\phi$ (e.g., GELU, tanh) can be applied to the state update at each chunk boundary or after the parallel update:
\begin{equation}
    \mathbf{S}_{[t,\,t+C-1]} \leftarrow \phi(\mathbf{S}_{[t,\,t+C-1]})
\end{equation}
This is controlled by a flag in the implementation and can improve model capacity and stability.

\paragraph{Expressivity and Efficiency.}

By increasing the order $n$, DeltaProduct interpolates between the original DeltaNet ($n=1$) and a dense state transition. Notably, DeltaProduct of order $n=2$ achieves a level of expressivity comparable to the Titans model and can solve algorithmic problems beyond the class $\mathrm{TC}^0$ (such as parity)~\cite{siems2025deltaproduct, merrill2024illusion}. This is in contrast to diagonal or single-rank updates, which are fundamentally limited in this regard. DeltaProduct thus provides a tunable trade-off between efficiency and expressivity, while retaining the hardware efficiency of the chunkwise parallel algorithm.

\subsection{Integration with Pretrained Models}

Our framework modifies pretrained Transformer models by injecting linearized attention and memory augmentation modules. The procedure consists of four steps: (i) \textbf{Identification of target modules:} utilities such as \texttt{get\_tptt\_model} assist in locating the attention blocks to be replaced; (ii) \textbf{Modification of attention layers:} selected blocks are substituted with the \texttt{LiZAttention} module, which combines standard and linear attention with optional weight sharing among Q, K, V, and O projections and integrates the MaG mechanism; (iii) \textbf{Parameter-efficient fine-tuning:} the modified model is adapted using approaches such as LoRA (via \texttt{get\_peft\_model}), ensuring compatibility with pretrained weights while limiting the number of trainable parameters; (iv) \textbf{Persistence and reuse:} dedicated functions (\texttt{save\_tptt\_safetensors}, \texttt{load\_tptt\_safetensors}) allow models to be stored and reloaded efficiently. This pipeline allows any causal pretrained LLM to be transformed into a memory-augmented, long-context capable architecture with minimal architectural disruption.

\subsection{LiZAttention Module}

The \texttt{LiZAttention} operator represents the core architectural modification introduced by TPTT. It implements a dual-path attention mechanism that combines (i) standard softmax attention to retain expressivity and (ii) a computationally efficient linearized variant that scales to longer sequences. A gating coefficient $\alpha$, learned during fine-tuning, adaptively reweights these contributions at inference time.

To further minimize overhead, projection matrices may optionally share parameters across the two branches, ensuring that Q, K, V, and O projections are reused between softmax and linearized attention. This reduces memory footprint without substantially impacting representational capacity.

\begin{algorithm}[ht]
\caption{LiZAttention Forward Pass}
\label{alg:lizattention}
\begin{algorithmic}[1]
\Require $\text{hidden states} \in \mathbb{R}^{B \times L \times D}$ \Comment{Batch size $B$, length $L$, hidden dimension $D$}
\Require $\text{mask}$ \Comment{Causal/padding attention mask}
\State \textbf{Projection step:} Compute queries $q$, keys $k$, values $v$ via learned projections.
\State \textbf{Masking:} Apply $\text{mask}$ to enforce causality and handle padding.
\State \textbf{Linear attention:} Compute output $o_{\mathrm{lin}}[t]$ using a kernelized feature map $\phi$:
\begin{equation*}
o_{\mathrm{lin}}[t] = \frac{\sum_{i=1}^{t} \phi(q_t)^\top \phi(k_i) v_i}{\sum_{i=1}^{t} \phi(q_t)^\top \phi(k_i)}.
\end{equation*}
Intermediate states are cached for reuse across forward passes.
\State \textbf{Softmax attention:} Compute self-attention output (ex: LlamaAttention):
\begin{equation*}
o_{\mathrm{base}} = \text{Softmax}\left(\frac{q k^\top}{\sqrt{d_h}}\right) v.
\end{equation*}
\State \textbf{Gating (MaG):} Combine both outputs using a learnable coefficient $\alpha \in [0,1]$,
\begin{equation*}
o = \alpha \, o_{\mathrm{lin}} + (1-\alpha) \, o_{\mathrm{base}}.
\end{equation*}
\State \Return $o \in \mathbb{R}^{B \times L \times D}$
\end{algorithmic}
\end{algorithm}

\subsubsection{Efficient Internal Memory Management}

In practice, \texttt{LiZAttention} maintains a bounded cache of intermediate representations rather than storing the entire history. Specifically, for the derivative-based virtual token expansion, the cache is limited to the most recent $n$ tokens preceding the current step, where $n = n_h$ corresponds to the order of the Householder expansion. This design allows the mechanism to exploit local temporal structure while avoiding unbounded growth in memory requirements. 

The caching strategy thus ensures that attention computations scale linearly with sequence length, as only a fixed-size window of past states is reused instead of recomputed. To further balance expressivity and efficiency, the cached states may optionally be compressed or unlinear (e.g., via GELU) before reuse. This combination enables \texttt{LiZAttention} to capture extended dependencies beyond the immediate context, while retaining predictable and tractable memory usage.

\section{Training Procedure}\label{sec:training}

\subsection{Parameter-Efficient Fine-Tuning with LoRA}

To adapt pretrained Transformer models augmented with TPTT components, we employ Low-Rank Adaptation (LoRA)~\cite{hu2022lora, lora_hf}, a parameter-efficient fine-tuning method. LoRA introduces trainable low-rank matrices into selected projection layers while freezing the original pretrained weights, thereby substantially reducing the number of trainable parameters and lowering memory consumption during training. In our experiments, unless specified otherwise, LoRA is configured with rank 8, a scaling factor \(\alpha=16\), and a dropout rate of 0.05. The adaptation is applied to primary projection modules, namely \texttt{[q\_proj, k\_proj, v\_proj, o\_proj]} for Llama and Mistral architectures, and \texttt{[qkv\_proj, out\_proj]} for OpenELM. For comparison, we also consider a setting where only the projection matrices are made trainable without applying LoRA updates (denoted LoRA\(-\)).

\subsection{Dynamic Memory as Gate Scheduling (LiZA Callback)}

An important component of our training procedure is the LiZA callback, which dynamically controls the Memory as Gate (MaG) weighting parameter \(\alpha\) during training. We consider three main scheduling strategies for \(\alpha\), each providing different mechanisms to balance the contributions of softmax and linearized attention:

\begin{itemize}
    \item \textbf{Constant scheduling:} \(\alpha\) remains fixed at a preset value throughout training, such as 0.125, 0.5, or 1.0. This maintains a steady fixed balance between the two attention mechanisms.
    \item \textbf{Gradual scheduling:} \(\alpha\) starts from a small initial value (e.g., 0.01) and increases linearly to a target value (commonly 0.5) over a predefined number of initial training steps (e.g., first 100 steps). This enables a smooth transition from predominantly softmax attention to more reliance on linearized attention, potentially enhancing training stability and convergence.
    \item \textbf{Cyclic scheduling:} \(\alpha\) cycles periodically through a set of discrete values (e.g., 0, 0.5, and 1.0) during training, allowing the model to alternate phase-wise between different attentional emphases.
\end{itemize}

These three cases provide complementary approaches to control the trade-off between efficiency and expressivity during fine-tuning.

\subsection{Experimental Variants}

To comprehensively evaluate the effects of different architectural and training choices, we tested several TPTT variants that differ across multiple dimensions:

\begin{itemize}
    \item \textbf{Memory gating coefficients \(\beta\):} The gating scalars modulating the keys and values can be based on either \(\beta(k)\), \(\beta(v)\), or the elementwise product \(\beta(k) \times \beta(v)\). This modulation controls the relative importance of each token's contribution and influences the attention dynamics.
    \item \textbf{Memory as Gate (MaG) combination method:} The final attention output can be combined either through a simple additive mixture of linearized and softmax attention outputs or via a nonlinear cross-gate interaction, i.e., elementwise multiplication of the weighted outputs, allowing richer expressivity.
    \item \textbf{Number of Householder updates (\(n_h\)):} We explored simple Delta Rule-based variants with \(n_h=1\) reflecting a single Householder transformation, as well as more expressive DeltaProduct variants with \(n_h=2\) enabling a product of multiple Householder reflections for richer state transitions.
    \item \textbf{Virtual token expansion strategies:} These include the “derivative trick” (\(\Delta\)), which computes finite difference approximations of input sequences, and the “rotary trick” (\(\Theta\)), which applies rotational embeddings. Variants combining both (\(\Delta,\Theta\)) were also tested, enhancing the feature representations sent into the memory updates.
    \item \textbf{Nonlinearity between chunks:} Some models include a nonlinear activation function such as GELU applied after chunk-wise state updates to improve model capacity and stabilization.
    \item \textbf{Fine-tuning approach:} Variants trained either with parameter-efficient Low-Rank Adaptation (LoRA) or without LoRA (denoted LoRA\(-\)) to assess adaptability and parameter efficiency impacts.
    \item \textbf{MaG weight scheduling \(\alpha\):} We compared constant \(\alpha\) values, a gradual linear increase schedule, and cyclic schedules where \(\alpha\) periodically varies among discrete levels to probe different balances between softmax and linearized attention during training.
\end{itemize}

This diversity captures important axes influencing model behavior in long-context attention and memory management. The following table summarizes these principal configurations, facilitating interpretation of related training results and curve labels, often formatted as \texttt{Model::LiZA(parameters)}, where “parameters” encode the above characteristics.

\begin{table}[ht]
\centering
\renewcommand{\arraystretch}{1.3}
\begin{tabular}{|l|p{11cm}|}
\hline
\textbf{Parameters Description} & \textbf{Details} \\
\hline
$k$, $+$, $n_h=1$, $\alpha=0.5$ & MaG gating using \(\beta(k)\) scalar, additive combination, single Householder (Delta Rule), constant \(\alpha=0.5\) \\
$k$, GELU, $n_h=1$, $\alpha=0.5$ & Gating with \(\beta(k)\), GELU activation between chunks, single Householder, constant \(\alpha=0.5\) \\
$\Delta$, $k$, $+$, $n_h=2$, $\alpha=0.5$ & Derivative trick, additive MaG, two Householder projections, constant \(\alpha=0.5\) \\
$v$, $+$, $n_h=1$, $\alpha=0.5$ & Gating with \(\beta(v)\), additive MaG, one Householder, constant \(\alpha=0.5\) \\
$(k,v)$, $+$, $n_h=1$, $\alpha=0.5$ & Gating with product \(\beta(k) \times \beta(v)\), additive MaG, one Householder, constant \(\alpha=0.5\) \\
$\Theta$, $k$, $+$, $n_h=2$, $\alpha=0.5$ & Rotary trick virtual tokens, gating \(\beta(k)\), additive MaG, two Householder, constant \(\alpha=0.5\) \\
$(\Delta,\Theta)$, $k$, $+$, $n_h=2$, $\alpha=0.5$ & Combination of derivative and rotary tricks, gating \(\beta(k)\), additive MaG, two Householder updates, constant \(\alpha=0.5\) \\
$(\Delta,\Theta)$, $k$, $+$, $n_h=2$, $\alpha=1.0$ & Same combined tricks, full reliance on linear attention (\(\alpha=1.0\)) \\
$\Delta$, $k$, $+$, $n_h=2$, $\alpha=0.5$, LoRA$-$ & As above without LoRA fine-tuning (frozen pretrained weights except projections), constant \(\alpha=0.5\) \\
$\Delta$, $k$, $\odot$, $n_h=2$, $\alpha=0.5$ & Derivative trick with cross-gate mixing (elementwise product), gating \(\beta(k)\), two Householder, constant \(\alpha=0.5\) \\
$\Delta$, $k$, $+$, $n_h=2$, $\alpha$ $\in [0,0.5]$ & Derivative trick, additive MaG, two Householder, gradual linear increase of \(\alpha\) from 0 to 0.5 during training \\
$\Delta$, $k$, $+$, $n_h=2$, $\alpha$ $\in \{0, 0.5, 1.0\}$ & Derivative trick, additive MaG, two Householder, cyclic \(\alpha\) switching among \{0, 0.5, 1.0\} during training \\
$k$, $+$, $n_h=1$, $\alpha=0.125$ & MaG gating \(\beta(k)\), additive, single Householder, constant low \(\alpha=0.125\) \\
$\Delta$, $k$, $+$, $n_h=2$, $\alpha=0.125$ & Derivative trick, additive MaG, two Householder, constant low \(\alpha=0.125\) \\
$\Delta$, $k$, $+$, $n_h=2$, $\alpha=1.0$ & Derivative trick, full linear attention reliance with constant \(\alpha=1.0\) \\
\hline
\end{tabular}
\caption{Summary of TPTT variants evaluated, indicating gating scalars (\(\beta(k)\), \(\beta(v)\), or their product), MaG combination method, virtual token expansions, Householder update order, nonlinearities, fine-tuning approach, and MaG weighting schedules.}
\label{tab:variant_parameters}
\end{table}

\section{Experiments and Results}

\subsection{Experimental Setup}

We evaluated the TPTT library on several pretrained language models with approximately 1 billion parameters, using the MMLU benchmark~\cite{hendrycks2020measuring} as the primary evaluation suite. Training was conducted on 500 samples from the \texttt{yahma/alpaca-cleaned} dataset~\cite{taori2023alpaca} for 5 epochs, with a maximum sequence length of 512 tokens, a batch size of 2, and a learning rate of $5 \times 10^{-4}$. Mixed precision training and gradient clipping at 1.0 were employed to optimize computational efficiency and stability. All experiments were performed on NVIDIA Tesla T4 GPUs (Kaggle platform). The trained models and detailed metrics are publicly available on the Hugging Face Model Hub\footnote{\url{https://huggingface.co/ffurfaro/}}, with full training logs accessible via Hugging Face TensorBoard.

\subsection{Training Results}

Table~\ref{tab:training-metrics} reports key training performance metrics for various pretrained models augmented with the TPTT framework. Across different architectures, including variants of Llama, OpenELM, Gemma, Qwen, OLMo, and Mistral, TPTT achieves consistently low final training losses, indicating effective optimization of the augmented models within relatively short training times.

\begin{table}[h!]
    \centering
    \resizebox{\textwidth}{!}{
    \begin{tabular}{|l|c|c|c|c|c|c|l|}
    \hline
    Model & Loss & Training Time (s) & Samples/s & Steps/s & Total FLOPs & Gradient Norm & Refs. \\
    \hline
    Titans-Llama-3.2-1B           & 1.375   & 1654.1  & 1.51  & 0.254 & $5.62 \times 10^{15}$   & 2.68 & \cite{touvron2023llama} \\
    Titans-OpenELM-1\_1B          & 1.3188  & 1651.1  & 1.51  & 0.254 & $5.85 \times 10^{15}$   & 0.704 & \cite{mehta2024openelm} \\
    Titans-Qwen2.5-1.5B           & 1.2568  & 1900.6  & 1.31  & 0.221 & $7.56 \times 10^{15}$   & 1.99 & \cite{bai2023qwen} \\
    Titans-OLMo-1B-hf             & 1.3068  & 1585.2  & 1.58  & 0.265 & $6.20 \times 10^{15}$   & 3.12 & \cite{groeneveld2024olmo} \\
    Titans-v2-OLMoE-1B-7B-0924    & 1.690   & 8080.4  & 0.247 & 0.195 & $5.31 \times 10^{15}$   & 2.04 & \cite{muennighoff2024olmoe} \\
    Titans-v2-Mistral-7B-v0.3     & 1.060   & 11512.6 & 0.174 & 0.089 & $5.57 \times 10^{15}$   & 5.15 & \cite{jiang2023mistral7b} \\
    Titanesque-gemma-3-270m       & 2.21    & 1318.9  & 1.96  & 0.495 & $2.01 \times 10^{15}$   & 1.81 & \cite{team2025gemma} \\
    \hline
    \end{tabular}
    }
    \caption{Training performance of Titans-based models with TPTT.}
    \label{tab:training-metrics}
\end{table}

Throughput metrics, expressed in samples per second and steps per second, reflect efficient training dynamics despite the additional linearized attention and memory gating mechanisms integrated by TPTT. For example, models such as Titans-Llama-3.2-1B and Titans-OpenELM-1\_1B maintain sample processing rates above 1.5 samples/s, demonstrating modest performance overhead relative to their base architectures.

Gradient norms remain stable across experiments, suggesting that the inclusion of linearized attention and the Memory as Gate (MaG) modules does not introduce training instability or gradient explosion issues. The total floating-point operations (FLOPs) required are largely manageable, indicating that TPTT’s enhancements induce only moderate computational cost increases, which are justified by the potential gains in long-context handling capabilities.

Notably, larger models such as Titans-v2-Mistral-7B-v0.3 display higher training times and reduced throughput, which is consistent with their increased size and complexity. However, the framework continues to support practical training within resource constraints (e.g., using NVIDIA Tesla T4 GPUs). Similarly, smaller models like Titanesque-gemma-3-270m exhibit faster training with higher throughput, illustrating the scalability of TPTT across a wide range of model scales.

Overall, these training results suggest that TPTT can enhance diverse pretrained Transformer models with advanced attention and memory mechanisms while maintaining efficient and stable training properties.

\subsection{Analysis of Training Curves}

\begin{figure}[ht]
    \centering
    \includegraphics[width=1.0\linewidth]{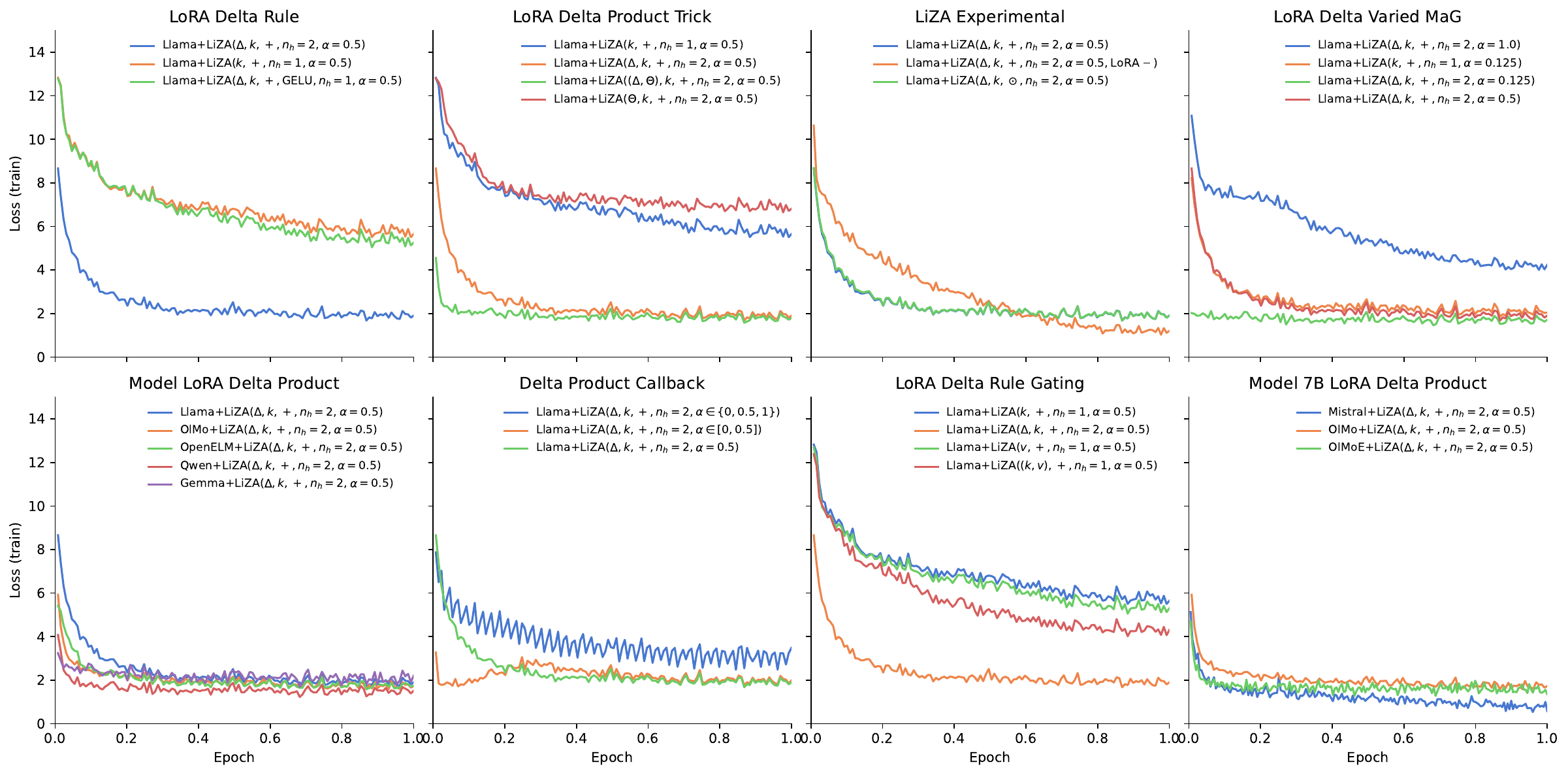}
    \caption{Experimental training curves for TPTT models on the Alpaca dataset. The plots show training loss over training steps, demonstrating stable convergence and effective learning dynamics across different architectures.}
    \label{fig:training_curve}
\end{figure}

Figure~\ref{fig:training_curve} illustrates training loss trajectories for multiple TPTT configurations on the Alpaca dataset. The experiments span different model architectures, memory update methods, gating strategies, and schedules for the Memory as Gate (MaG) parameter \(\alpha\).

\paragraph{LoRA Delta Rule Variants}

The training curves associated with \texttt{Llama+LiZA(k, +, nh=1, 0.5)} and \texttt{Llama+LiZA(k, +, GELU, nh=1, 0.5)} show similar behavior, indicating that applying a GELU nonlinearity between chunks does not yield a clearly measurable impact on convergence speed or final loss. Models using two Householder updates (\(n_h=2\)), such as \texttt{Llama+LiZA(k, +, nh=2, 0.5)}, tend to converge faster and reach lower training losses within one epoch compared to the single update variants. This suggests that increasing the complexity of memory updates may improve learning efficiency.

\paragraph{LoRA Delta Product Variants}

Among DeltaProduct variants, the one employing only the rotary trick 
\texttt{Llama+LiZA(}\(\Theta\)\texttt{, k, +, nh=2, 0.5)} exhibits slower convergence than the variant using the derivative trick \texttt{Llama+LiZA(}\(\Delta\)\texttt{, k, +, nh=2, 0.5)}. This finding contrasts with the expectation that rotary embeddings enhance model capacity, implying that the rotary trick alone may not improve performance in this context. The combination of derivative and rotary tricks appears to produce the most favorable convergence speed and lowest training loss—albeit with increased computational cost—indicating the potential complementary benefits of these expansions.

\paragraph{Effect of Fine-Tuning and Special Modes}

The comparison between models trained with and without Low-Rank Adaptation (LoRA) reveals that while LoRA facilitates faster initial convergence, models trained without LoRA achieve marginally better final loss values. The impact of applying cross-gate $\odot$ mixing appears limited in this experimental setting.

\paragraph{Impact of MaG Scheduling}

Analyses of \(\alpha\) scheduling reveal distinct behaviors when separating constant from dynamic schedules:

\begin{itemize}
    \item Under constant \(\alpha\), lower values such as 0.125 correspond to relatively fast and stable convergence, even with simpler Delta Rule updates (\(n_h=1\)). \textbf{Surprisingly}, a constant \(\alpha\) of 1.0, representing exclusive reliance on linear attention, converges steadily despite the lack of softmax attention, which may indicate that the model can adapt to fully linearized attention given sufficient training.
    \item Regarding dynamic schedules, gradual increase of \(\alpha\) from 0.01 to 0.5 does not demonstrate clear advantages over a fixed \(\alpha = 0.5\). Cyclic schedules typically yield slightly slower convergence overall but enable faster training of models with \(\alpha=1.0\), potentially offering a useful trade-off between convergence speed and generalization capability.
\end{itemize}

\paragraph{Gating Strategy Comparisons}

Comparisons among gating on keys \(\beta(k)\), values \(\beta(v)\), or their product \(\beta(k) \times \beta(v)\), reveal minimal differences between gating on keys or values individually. Combined gating provides a modest improvement in convergence speed but remains less impactful than increasing the number of Householder updates.

\paragraph{Generalization Across Architectures and Model Sizes}

Training loss behaviors are consistent across diverse pretrained architectures including \texttt{Olmo}, \texttt{OpenELM}, \texttt{Qwen}, \texttt{Gemma}, and larger models like \texttt{Mistral} and \texttt{OlmoE}, all evaluated with DeltaProduct and \(\alpha=0.5\). Larger models generally attain lower loss values, consistent with their increased capacity, while smaller models converge less rapidly. These results suggest the TPTT framework is adaptable and stable across a range of model sizes and architectures.

\subsection{Evaluation Metrics and Benchmark Results}

For evaluation, we focus on metrics standard in LLM and QA benchmarking (\textit{Measuring Massive Multitask Language Understanding}): Exact Match (EM), Partial Exact Match (PEM), and Partial Quasi Exact Match (PQEM). These metrics respectively measure strict correctness, partial overlap, and quasi-exactness between model outputs and ground truth answers, providing a nuanced view of model performance~\cite{hendrycks2020measuring}.

Table~\ref{tab:mmlu-results} presents the MMLU benchmark results in the one-shot setting. TPTT models, and especially Titans-Llama-3.2-1B, consistently outperform their base counterparts in EM, with better PEM and PQEM scores. This shows the benefit of integrating linearized attention and internal memory mechanisms for complex language understanding tasks.

\begin{table}[h!]
    \centering
    \renewcommand{\arraystretch}{1.2}
    \begin{tabular}{|l|c|c|c|}
    \hline
    Model & EM & PEM $\pm$ Std & PQEM $\pm$ Std \\
    \hline
    Titans-Llama-3.2-1B & 0.246 $\pm$ 0.128 & 0.265 $\pm$ 0.134 & 0.477 $\pm$ 0.157 \\
    Llama-3.2-1B        & 0.007 $\pm$ 0.006 & 0.311 $\pm$ 0.141 & 0.472 $\pm$ 0.153 \\
    \hline
    Titans-Qwen2.5-1.5B & 0.000              & 0.500 $\pm$ 0.150 & 0.583 $\pm$ 0.144 \\
    Qwen2.5-1.5B        & 0.000              & 0.598 $\pm$ 0.142 & 0.690 $\pm$ 0.129 \\
    \hline
    Titans-OLMo-1B-hf   & 0.000              & 0.261 $\pm$ 0.131 & 0.465 $\pm$ 0.154 \\
    OLMo-1B-hf          & 0.000              & 0.233 $\pm$ 0.130 & 0.425 $\pm$ 0.153 \\
    \hline
    \end{tabular}
    \caption{One-shot MMLU benchmark results with mean $\pm$ standard deviation. 
    Values reported as 0.000 indicate that no successful cases were observed across all runs. 
    Each pair compares a Titans model with its corresponding base model.}
    \label{tab:mmlu-results}
\end{table}

Compared to recent state-of-the-art methods such as Mamba~\cite{gu2023mamba}, LoLCat~\cite{zhang2024lolcats}, and Liger~\cite{lan2025liger}, TPTT stands out by enabling the transformation of existing pretrained models without full retraining, while maintaining good benchmark performance. The observed improvements in EM and better PEM/PQEM scores highlight the effectiveness of TPTT's linearized attention and memory augmentation for efficient and robust LLM adaptation.

\section{Discussion and Conclusion}

This work presents TPTT, a framework designed to enhance pretrained Transformer models by integrating efficient linearized attention mechanisms together with internal memory augmentation. Leveraging parameter-efficient fine-tuning via LoRA~\cite{hu2022lora}, TPTT facilitates effective adaptation of large language models (LLMs) to long-context tasks without requiring full retraining. Experimental evaluations on the MMLU benchmark~\cite{hendrycks2020measuring} suggest improvements in both computational efficiency and accuracy, supported by statistical analyses and favorable comparisons to recent state-of-the-art methods~\cite{gu2023mamba, zhang2024lolcats, lan2025liger}.

\paragraph{Summary of Findings}  
Our results indicate that TPTT enables efficient and effective augmentation of pretrained Transformers with memory-augmented attention. In particular, higher-order DeltaProduct updates combined with diverse token expansion strategies contribute to improved convergence speed and model capacity. While LoRA fine-tuning accelerates early training phases, full fine-tuning can sometimes yield superior final performance. Fixed schedules for gating parameters generally suffice, and dynamic weighting strategies produce mixed benefits depending on the scenario. Importantly, these outcomes generalize well across several pretrained model families, supporting the framework’s broad applicability. These findings remain preliminary and should encourage further community replication, testing, and extension. Taken together, these observations suggest that:

\begin{itemize}
    \item Utilizing DeltaProduct with multiple Householder updates (\(n_h=2\)) may enhance convergence speed and final training performance compared to simpler Delta Rule variants.
    \item Token expansion via combined derivative and rotary tricks could contribute additional capacity, though with computational trade-offs.
    \item LoRA fine-tuning facilitates faster initial convergence; however, full fine-tuning may yield improved final metrics in certain cases.
    \item Fixed MaG weightings around 0.5 appear sufficient for effective training, with dynamic adjustments providing no consistent benefits under the evaluated conditions.
    \item Gating strategies have limited impact compared to the model’s memory update complexity.
    \item The training dynamics generalize well across different pretrained models, supporting the practical applicability of the proposed approach.
\end{itemize}

These findings are preliminary and subject to further validation over larger datasets, longer training schedules, and diverse benchmarks. They indicate that TPTT may offer a tractable approach to incorporate efficient, memory-augmented mechanisms into pretrained language models for long-context tasks.

\paragraph{Practical Implications}  
TPTT provides a scalable, practical approach to deploying high-performance LLMs in environments with limited computational and memory resources. By combining linearized attention with internal memory augmentation, the framework reduces inference costs while potentially enhancing accuracy on tasks involving long-range dependencies. Its seamless integration with popular frameworks and reliance on LoRA fine-tuning help lower barriers for real-world adoption and rapid domain adaptation.

\paragraph{Limitations}  
Current evaluations focus primarily on models ranging from approximately 1 to 7 billion parameters. Extending TPTT to larger-scale architectures and a wider variety of downstream applications poses challenges, including increased tuning complexity, memory management difficulties, and computational overhead. Further, more extensive empirical validation across diverse benchmarks and real-world datasets is necessary to more comprehensively assess robustness, stability, and practical utility. Additionally, while core linearized attention and MaG mechanisms are evaluated, only instruction-tuning tasks are considered; More complex domains, including reasoning, causal inference, and interactive settings like reinforcement learning~\cite{shao2024deepseekmath}, remain unexplored.

\paragraph{Future Directions}  
Future work will concentrate on optimizing the integration and efficiency of linearized attention and memory modules, exploring advanced internal memory mechanisms~\cite{behrouz2025s}, and broadening evaluations to larger models and more diverse task domains. Investigating hybrid methods that combine linearized attention with complementary efficiency-oriented techniques, such as quantization, pruning, or sparse attention, could yield further gains. Moreover, the bidirectional linear attention capability present in TPTT, currently unused in this study, opens promising avenues for applications in text-to-text understanding ~\cite{devlin2019bert}, multi-modal tasks like image understanding ~\cite{dosovitskiy2020image}, or even diffusion LLMs~\cite{ye2025dream}. Comprehensive benchmarking on these fronts will clarify the full potential of TPTT in future AI systems.

In summary, TPTT represents a practical, scalable, and effective software library for upgrading pretrained Transformer models with memory-augmented attention. Supported by strong empirical results, it offers promising implications for advancing efficient language modeling tailored for long-context and resource-constrained scenarios.

\bibliographystyle{plain}
\bibliography{refs}

\end{document}